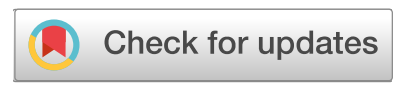

OPEN

# Training convolutional neural networks with the Forward–Forward Algorithm

Riccardo Scodellaro[1✉], Ajinkya Kulkarni[1], Frauke Alves[1,2,3] & Matthias Schröter[3,4]

**Recent successes in image analysis with deep neural networks are achieved almost exclusively with Convolutional Neural Networks (CNNs), typically trained using the backpropagation (BP) algorithm. In a 2022 preprint, Geoffrey Hinton proposed the Forward–Forward (FF) algorithm as a biologically inspired alternative, where positive and negative examples are jointly presented to the network and training is guided by a locally defined goodness function. Here, we extend the FF paradigm to CNNs. We introduce two spatially extended labeling strategies, based on Fourier patterns and morphological transformations, that enable convolutional layers to access label information across all spatial positions. On CIFAR10, we show that deeper FF-trained CNNs can be optimized successfully and that morphology-based labels prevent shortcut solutions on dataset with more complex and fine features. On CIFAR100, carefully designed label sets scale effectively to 100 classes. Class Activation Maps reveal that FF-trained CNNs learn meaningful and complementary features across layers. Together, these results demonstrate that FF training is feasible beyond fully connected networks, provide new insights into its learning dynamics and stability, and highlight its potential for neuromorphic computing and biologically inspired learning.**

Machine learning using deep neural networks (DNN) continues to transform human life in areas as different as art (DALL-E, stable diffusion), medicine (Alpha-Fold), transport, or natural language models (ChatGPT, Gemini). Here, the adjective "deep" refers to the number of layers of artificial neurons, which can reach hundreds. Training these networks means shifting the weights that connect the layers from their initial random values to values that produce the correct predictions at the DNN output layer. This is achieved with the help of a loss function that computes the aggregate difference between the predicted output and the accurate results, which must be known for the training examples. The algorithm behind the training is some variant of gradient descent: in each round of training, each weight is shifted a bit into the direction minimizing the loss by using the derivative of the loss function with respect to that weight. Taking the derivative of a loss function with respect to a given weight is straightforward for a single output layer. Training the weights of the earlier layers in DNNs requires iterative gradient computation by applying the chain rule[1]. This process is called backpropagation (BP). Due to its importance, the term BP is also often used loosely to refer to the entire learning algorithm, including the gradient descent.

Backpropagation, respectively multi-layer gradient descent, has a number of downsides. First, it requires the storage of intermediate results. Depending on the optimizer, the memory consumption of BP is up to five times larger than the requirement to store the weights alone[2]. This becomes a problem when training large models on GPU cards with limited memory. Second, under the name neuromorphic computing, there is an ongoing search for hardware alternatives to CMOS semiconductors, driven by the desire to reduce power consumption and increase processing speed[3]. On these new hardware platforms, it is often impossible to implement an analog of BP, raising the need for an alternative training algorithm. Finally, evolution has clearly developed learning algorithms for neural networks such as our brain. However, those algorithms seem quite (but maybe not completely[4]) different from BP. Given the in general high performance of evolutionary solutions to problems, this raises the question whether deep learning could also benefit from biologically plausible alternatives to BP.

[1]Translational Molecular Imaging, Max Planck Institute for Multidisciplinary Sciences, Hermann-Rein Straße 3, 37075 Göttingen, Germany. [2]Department of Haematology and Medical Oncology, University Medical Center Göttingen, Robert Koch-Straße 40, 37075 Göttingen, Germany. [3]Institute for Diagnostic and Interventional Radiology, University Medical Center Göttingen, Robert Koch-Straße 40, 37075 Göttingen, Germany. [4]Max Planck Institute for Dynamics and Self-Organization, Am Faßberg 17, 37075 Göttingen, Germany. ✉email: riccardo.scodellaro@mpinat.mpg.de

Due to these limitations, there is an ongoing search for alternative training methods. Several approaches have been proposed over the years, which we summarize in the following Related Work section. Among the most recent proposals, a novel example of the use of local information collected in consecutive runs to update the weights is the Forward-Forward (FF) algorithm[5]. FF training combines two ideas. First, the weights of a given layer are updated using gradients of a locally defined goodness function, which is taken to be the sum of the squares of activities in that layer[5]. Second, labels are included in the training data, which allows neurons to learn them together. In order to understand which features of the data vote for a given label, half of the dataset is made up of labels combined with wrong images. For these negative data the weights are changed in order to minimize the goodness. In contrast, for the correctly labeled positive data the weights are modified to maximize the goodness. Both of these objectives can be achieved with a local gradient descent, without the need for BP. The term Forward-Forward refers to having two subsequent training steps, one with positive and one with negative data.

To date, applications of the Forward-Forward (FF) algorithm do not include convolutional neural networks (CNN)[6], despite CNNs being one of the most widely used architectures in image analysis. We close this gap by extending the FF paradigm to CNNs and testing it on three benchmark datasets: MNIST, CIFAR10 and CIFAR100. A key innovation of our work is the use of two novel spatially-extended labeling techniques, which ensures that label information remains accessible across all spatial positions and convolutional filters also in the presence of more complex and fine features. This enables CNN training under the FF framework on datasets of varying complexity. In addition, we provide a systematic investigation of hyperparameter tuning for FF-CNNs and demonstrate, through explainable AI methods such as Class Activation Maps (CAMs), that FF-trained CNNs learn meaningful and complementary features. Our results show that with the appropriate settings, FF-trained CNNs achieve promising accuracy values, while offering new opportunities for biologically inspired learning and neuromorphic hardware implementation.

## Related works

A wide range of alternatives to BP have been proposed, aiming to address its limitations in terms of memory requirements, biological plausibility, and hardware compatibility. These methods can be grouped into several families, which we briefly review in the following.

Neural networks trained with variants of the locally acting Hebbian learning rule (neurons that fire together wire together) have been shown to be competitive with BP[7,8]. In addition to standard Hebbian formulations, Contrastive Hebbian Learning (CHL)[9] has been proposed as a biologically motivated alternative, where learning is driven by the difference between free and clamped network states. These approaches are conceptually close to the idea of local learning rules, and in some cases have been shown to approximate backpropagation in layered architectures[10]. Another approach, Equilibrium Propagation[11] is a learning framework for energy-based models with symmetric connections between neurons. Equilibrium Propagation and related energy-based methods bridge the gap between Hopfield-like models and gradient descent, and remain one of the most actively investigated biologically plausible frameworks[12]. Layer-wise Feedback Propagation[13] removes the need for a gradient computation by replacing the objective of reducing the loss with computing a reward signal from the network output and propagating that signal backward into the net. In contrast, the most conservative approach is to keep gradient descent, but to replace the required gradients with an estimate computed from the difference in loss of two forward passes with slightly modified weights. While the naive version of this approach, labeled zeroth order optimization, can be expected to be extremely inefficient, modern variants seem to be competitive[14–17]. A third category of algorithms maintains the idea of updating the weights using derivatives of some signal, which involves the difference between the present state of the network and the target state. However, it relaxes the requirement to backpropagate that signal from the output layer towards the earlier layers. This either can be done by using exclusively an output-derived error signal for training each intermediate layer[18,19], Direct Feedback alignment, Feed-Forward with delayed Feedback (F3) or by training each layer with locally available information collected in two consecutive forward passes. An example of the latter is the "Present the Error to Perturb the Input To modulate Activity technique" (PEPITA), which performs the second forward pass with the sum of the input signal used in the first pass and some random projection of the error signal from that pass[20,21]. Another biologically inspired family of learning rules are similarity matching algorithms, which formulate learning as the optimization of similarity between input and output representations[22]. These methods emphasize unsupervised feature discovery and highlight the diversity of gradient-free optimization strategies.

Within this broad landscape, FF can be understood as part of the family of local learning rules. When generalizing this training method to multilayer networks, it is important to ensure that each subsequent layer needs to do more than just measure the length of the activity vector of the previous. This is achieved using layer normalization[23], which normalizes the activity vector for each sample. This is best summarized by Hinton[5]: "the activity vector in the first hidden layer has a length and an orientation. The length is used to define the goodness for that layer and only the orientation is passed to the next layer.". There are two ways of using a FF trained network for inference. First, we can simultaneously train a linear classifier using the activities of the neurons in the different layers as input. Alternatively, we can create multiple copies of the dataset considered and combine each copy with one of the possible labels. The correct label is then the one with the largest goodness during its forward pass. Note that this approach multiplies the amount of computation required for inference by a factor equal to the number of labels.

Given the repute of the proposer, it is not surprising that the FF algorithm has inspired a number of groups to suggest modified and adapted versions. Examples include the combination with a generative model[24], multiple convolutional blocks (not trained with FF)[25], or extending the FF training to graph neural networks[26] and spiking neural networks[27]. In line with the motivation for neuromorphic computation to go beyond BP, the FF algorithm has also been used to train optical neural networks[28] and microcontroller units with low computational resources[29]. Another line of research tries to improve FF by modifying the way goodness is computed[30–32], by understanding the sparsity of activation in FF trained networks[33,34], or by exploring the capabilities of FF for self-supervised learning[35]. There has been less activity in terms of practical applications of the FF algorithm. In particular, in the classification of real-world images, FF has been observed to perform worse than BP[36] or has to be combined with BP to achieve satisfying results[37]. The first study applying the FF algorithm in a clinical imaging problem showed with explainable AI that the parallel use of BP and FF training unveils a wider set of information[38].

## Methods

This section will first discuss our new techniques for labeling the positive and negative datasets, before explaining our implementation of the FF algorithm in detail.

### Spatially-extended labeling

Fully connected DNNs establish connections between each pixel within an image and with each neuron in the next layer. The size of images is often of the order of megapixels and typical layers have hundreds of neurons. This results in a number of weights that need to be trained that far exceed the information available in typical training datasets. In contrast, convolutional layers use small filter kernels, typically sized in the range of 3 by 3 to 7 by 7 pixels, which are applied to all possible positions of the much larger input picture. In this way, each filter creates a new processed version of the input image. Because the training algorithm only needs to learn the weights in these filter kernels, it is possible to apply hundreds of these kernels in parallel and still have orders of magnitudes less parameters to train than in a fully connected network. FF training requires the labels to be added to the input images. Hinton achieves this with an one-hot encoding[5] , in which the label information is restricted to the first 10 pixels in the upper left region of each image. Figure 1a,b give an example of this technique.

However, convolutional layers will not work with this one-hot encoding because, for most of the possible positions of the filter kernel, these labels are not part of the input. For CNNs, it is imperative that the label information is spatially present throughout the entire image, ensuring that it is captured by each possible filter position. Moreover, this spatial labeling needs to be homogeneous; concatenated random patterns[30], will also not allow for arbitrary filter positions. Here, we introduced a spatial extended labeling approach involving superposition of the training dataset image with a second image of identical dimensions. We proposed two different alternatives to create the labels. The first one based on the creation of Fourier waves, where each possible label consists of a gray-value wave with a distinct configuration of frequency, phase, and orientation. The second one assigns each label a unique deterministic set of morphological transformations applied to the input image. This approach is advantageous for more complex datasets, as it forces the network to focus on morphological features rather than exploiting the simpler label-related patterns as shortcuts for distinguishing positive from negative images. Moreover, it does not rely on image dimensions or network hyperparameters and easily handle datasets characterized by many classes. More details are provided in the Appendix A of the Supplementary Materials. As indicated in Fig. 1c,d, both the positive and negative datasets were obtained by harnessing this methodology. After superposition of the label, the images were normalized to the range [0,1]. The relative contribution K of the label pattern to the total intensity of the image is a hyperparameter, whose influence was investigated in Appendix C of the Supplementary Materials. Note that we chose our negative labels randomly, not specifically hard based on another forward pass, as already suggested[5].

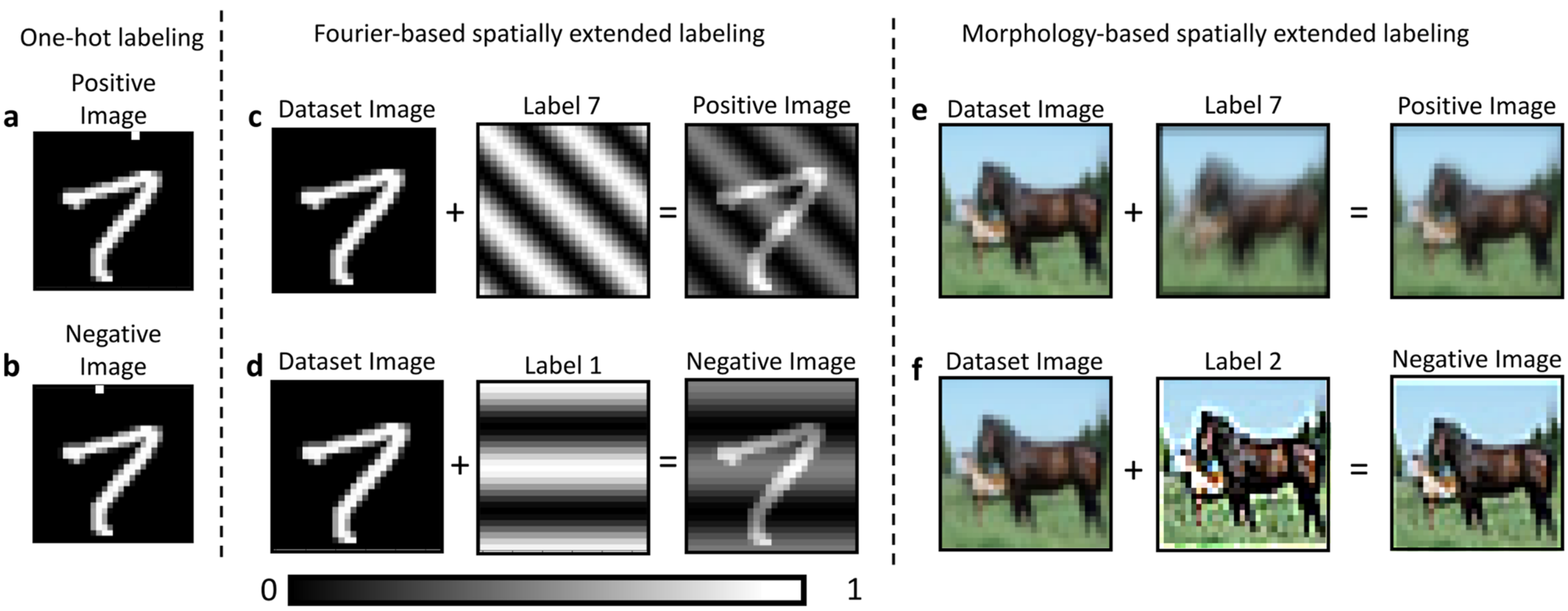

**Fig. 1**. Spatially-extended labels are present in the entire image, while one-hot encoding is confined to the upper-left area. For the FF training we need two datasets, which both add labels to the images. The top row describes the creation of the positive dataset, where the example image is correctly labeled. The bottom row displays an example of the negative dataset where the image is combined with a false label which was randomly chosen from the other possible ones. We display the three ways of adding the label. (**a**) and (**b**) describe the one-hot encoding used by Hinton: the first pixels in the top row of the image are used as indicators. In this example, the column number of the single pixels set to 1 corresponds to the target value. (**c**) and (**d**) describe the Fourier-based technique we propose. Each label corresponds to an image of the same size as the input, but with a characteristic gray value wave. The label is included into the image by pixel-wise addition. (**e**) and (**f**) show the morphology-based approach we propose. Each label is associated to a unique set of transformations which affect the image morphology, forcing the network in focusing on the image features.

## Implementation of the learning algorithm

For the study of the MNIST dataset, we used a network architecture composed of three consecutive FF-trained convolutional layers. All three layers contained the same number of filter matrices, which is one of the hyperparameter we examined. We did not add any max pooling layers because we found that those decrease accuracy as described in Appendix C of the Supplementary Materials. The flow of data through the network is shown in Fig. 2. Briefly, each input image, combined with its positive and negative labels, is sequentially passed through the convolutional layers with layer normalization and ReLU activation. In each layer, the discrepancies between positive and negative activations are used to compute the layer-specific sigmoidal function $\sigma$ of the goodness, whose values contribute to the binary cross-entropy loss guiding weight updates. During inference, feature maps can be connected to a linear classification layer, or alternatively, the cumulative goodness values across all layers can be compared for each possible label, with the final prediction given by the label associated with the maximum goodness.

Providing more details, the goodness is defined as the sum of squared layer activations $y_i$, modified by subtracting a user-provided threshold $\theta$. Following the Hinton code[5], and as confirmed in the literature[32], we choose $\theta$ to be equal to the number of neurons N within that layer. While computing the loss, we have to account for our different objectives regarding positive and negative data such as:

$$loss_{\,layer} = \sigma\left( \sum_{i=1}^{N} \begin{cases} y_i^2 - \theta & \text{if positive data} \\ -y_i^2 + \theta & \text{if negative data} \end{cases} \right)$$

Note that we did not induce symmetry in our loss as already proposed[30]. Indeed, we took inspiration from other works[31], which found improved collaboration between layers by training them with a cumulative network loss, which was computed by adding the individual losses of the layer. Here, we excluded the loss of the first layer since it yielded better accuracy, as shown in Appendix B of Supplementary Materials. This also aligns with Hinton's exclusion of the first layer during the evaluation phase[5]. Indeed, the length of the first hidden activity vector already separates positive from negative data. By normalizing and discarding this information, subsequent layers are forced to rely on relative activity patterns and thus learn new, meaningful features. We follow the Hinton implementation[5] in two more aspects. First, we applied layer normalization between the individual layers. Layer normalization involves the application of the following transformation to each activation $y_i$[23] as:

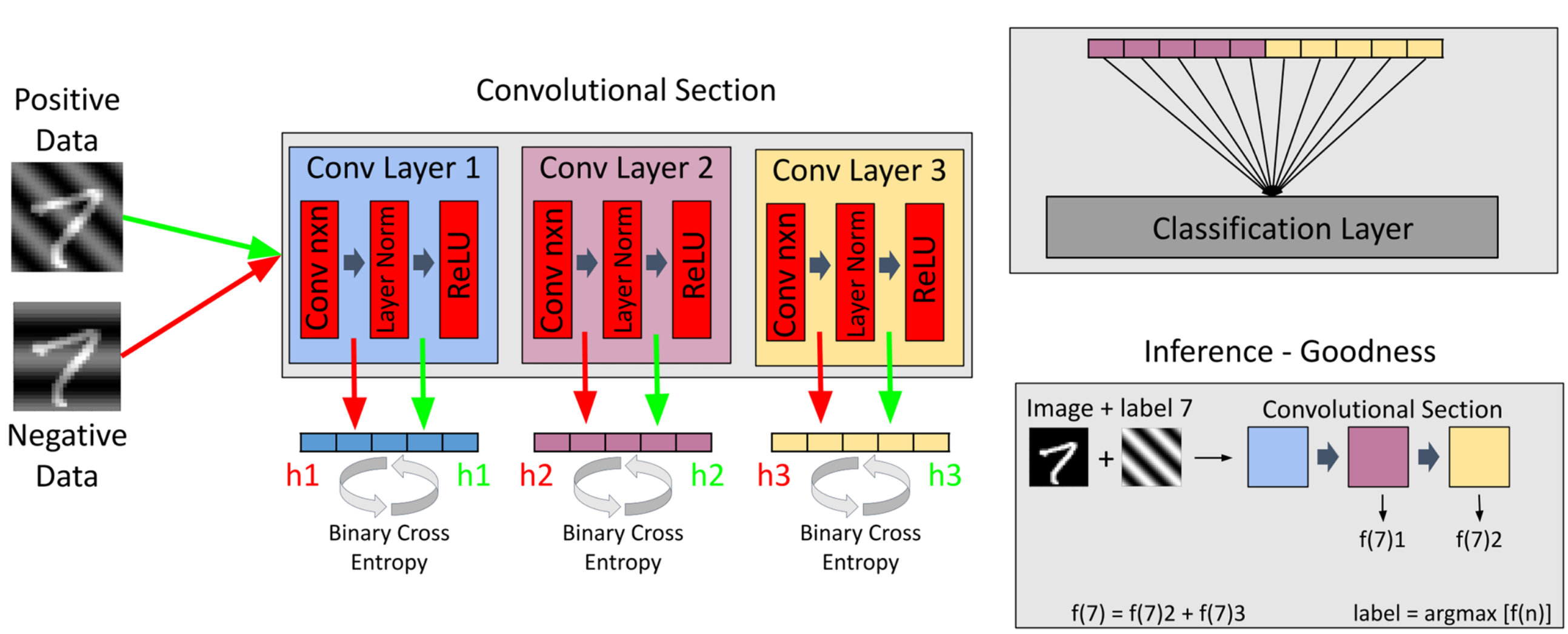

**Fig. 2.** Schematic overview of the FF-trained CNN applied to the MNIST dataset. Positive and negative samples are processed through three convolutional layers, each followed by layer normalization and ReLU activation. At every layer, the goodness function is computed using binary cross-entropy for both positive and negative samples. Final classification can then be performed either through a linear classifier or by evaluating the goodness scores across all labels.

$$y_{i,norm} = y_i \Bigg/ \left( \sqrt{\frac{\sum_{i=1}^{N} y_i^2}{N}} \right)$$

This ensures that each subsequent layer can only use the pattern, not the norm of the matrix formed by the activations of the previous layer. Second, the learning rate lr is modified halfway through the epochs by employing a linear cooldown:

$$lr(e) = \frac{2\ lr}{E}\,(1 + E - e)$$

where E represents the total number of epochs and e is the current epoch. To study the contribution of the individual layers, we defined a layer-based loss and accuracy which measures only the capability to discriminate between images of the positive and negative dataset. We interpreted the output of the Sigmoid function as a probability, where values greater than 0.5 indicate that the layer recognizes the image as belonging to the positive dataset. By comparing with the true assignment (positive or negative), we obtained a discrimination accuracy. Lastly, by using the probability to compute a binary cross entropy, we computed a layer-specific discrimination loss.

Considering inference, as shown in Fig. 2 and briefly explained in the data flow description, two options are viable: the linear classifier and the goodness evaluation. In the first case, H neurons of every layer (except the first) are fully-connected with an classification layer of N nodes, equal to the number of labels. The connecting weights, H times N in number, are trained by evaluating neuron activations using cross-entropy loss. This is the default method of inference used in this paper, unless otherwise mentioned. For inference with goodness evaluation, each image is exposed N times to the neural network, each time superimposed with another of the N possible labels, and the goodness parameter is computed for each label m. The image is then associated with $label_{correct}$, which is the label characterized by the highest goodness value, and is defined as:

$$label_{\ correct} = \text{argmax}([f_0, f_1, ..., f_8, f_9])$$

where for each associated label $m$, the goodness is expressed as $f_m = \sum_{i=1}^{H} y_i^2$, and $H$ is the number of all neurons, except those from the first layer.

Concerning the analysis of CIFAR10 and CIFAR100 datasets, we took advantage of some optimizations already discussed in the literature[39], designed for building deeper FF-trained networks. While keeping our spatially-extended labeling approach, we used a deeper CNN of 6 convolutional layers and we substituted the layer normalization with standard batch normalization. Details are provided in Appendix C of the Supplementary Materials.

### Hardware and software

The code for our FF trained CNN was implemented in Python using the PyTorch library[40]. The source code for an FF trained, fully connected DNN was used as the starting point[41]. All the displayed analysis was performed on a desktop workstation with an AMD Ryzen 9 5900X 12-Core Processor with 128 GB RAM, and an NVIDIA GeForce RTX 3080 GPU with dedicated 12 GB RAM. Preliminary tests were also performed on the supercomputer Grete at NHR-Nord@Göttingen as part of the NHR infrastructure.

## Results

We first report the configuration that achieved the highest accuracy on the MNIST dataset. A detailed exploration of the hyperparameter space leading to this optimal setting is provided in Appendix D of the Supplementary Materials. We also demonstrated via Class Activation Maps, an explainable AI method, that FF-trained CNNs leverage real features while performing classification tasks. We then extended our analysis to the more challenging CIFAR10 dataset. We also showed that training deeper CNNs (six convolutional layers) with the FF algorithm is feasible. Moreover, we highlighted that the choice of labeling strategy plays a critical role: if the labels are too simple compared to the morphological richness of the data, they dominate the learning process, prompting the network to exploit them as shortcuts for distinguishing positive and negative samples, rather than as meaningful guides for class discrimination. Finally, we evaluated the method on CIFAR100. Results show that the two proposed spatially-extended labeling strategies scale effectively, enabling discrimination across a larger number of classes, if diversity among the different labels is maximized. To ensure statistical robustness, all experiments were repeated five times with different random seeds, and we report mean values with standard deviations. Standard deviations smaller than the resolution of the figures are omitted for clarity.

### Performance of the optimized configuration on MNIST dataset

The hyperparameter optimization led to the following configuration for the FF trained CNN: three convolution layers of each 128 filters with a kernel dimension of 7x7 pixels. After training for 200 epochs with a batch size of 50 using the Adam optimizer with a learning rate of $5x10^{-5}$, and the label set 1 (please refer to the methods section for further details) with an intensity K of 35%, we obtained 99.20 ± 0.01% accuracy for the validation dataset and 99.16 ± 0.02% for the test dataset using the goodness approach for inference. Although having a significantly shorter run time, inference with the linear classifier approach provided slightly worse results, achieving accuracy values of 99.14 ± 0.02% and 99.00 ± 0.03% for validation and test datasets, respectively. For comparison, using again a three-layer CNN of constant size, but trained with BP, we obtained a validation accuracy of 99.13 ± 0.02%. Here, the search for optimal hyperparameters resulted in 16 filters of 5x5 pixels, Adam optimizer with a learning rate of $10^{-3}$, 200 epochs with batch size of 50. Figure 3 provides a more detailed picture of this comparison.

Figure 3a shows that the performance of the FF training increased monotonously with the number of filters per layer, while the accuracy of the BP training decreased slightly under the same conditions. The latter is most likely due to increasing overfitting. In fact, FF is characterized by slower convergence, whereas BP requires a smaller number of epochs to train the network[5]. This slower convergence comes with reduced memory requirements, which can reach 25–30%[42]. This could be due to the fact that FF could make a less efficient use of its number of trainable parameters: each hidden layer collects information only from the previous ones, while in BP, during backpropagation, each layer is also influenced by the following. This reflects a trade-off: indeed, FF gains from layer-local updates and reduced memory demands, which are advantageous in hardware-constrained settings. The confusion matrix in Fig. 3b provides insight into the classification performance, revealing that labels 4 and 9 exhibit the least accurate classifications (lower than 98.70%), while labels 1, 3 and 7 are characterized by the highest accuracy levels (higher than 99.20%). After 200 epochs, the discrimination loss of the FF trained network reached a plateau for all the considered layers (Fig. 3c). This convergence of the training is confirmed by a training run with 750 epochs, which results in no further substantial changes in accuracy. In addition, the accuracy value in the training data reached close to 100% (depicted in Fig. 3d with a green line) for 200 epochs. The discrimination accuracy values of the layers (red and blue lines in Fig. 3d) corroborated this result. They also hint at a slightly different interplay between the dynamics of layers 2 and 3, with layer 2 initially learning faster, but layer 3 achieving a higher faculty of discrimination in the long run.

## Explainable AI: class activation maps

Class Activation Maps (CAMs) are visual representations that highlight the regions of an input image that contribute the most to the prediction of a given label. CAMs are obtained by summing up the feature maps generated by the convolutional layers, each weighted with the corresponding weights associated with a specific label. The main underlying idea is that each feature map decodes specific spatial characteristics of the input image, and as a result, the weights quantify how much these characteristics contribute to the recognition of the target class. In BP trained CNNs, CAMs are typically obtained by applying a global average pooling layer after the last convolutional layer, followed by using the weights connecting this layer to a Softmax activation output layer. Here, we trained a CNN with FF and a linear classifier for inference. The weights of the linear classifier connecting the individual pixels in our feature map with the set of ten output neurons were exactly the weights we needed to assess the role of the corresponding pixel for a given prediction. Figure 4 provides four examples of CAMs of correctly identified images, obtained during both FF and BP-based trainings with MNIST dataset.

For digit 1, (Fig. 4a,e,i) the entire vertical shape contributed to the correct inference. In contrast, for the digit 2 of Fig. 4b three distinct areas (the upper, bottom-left and bottom right parts of the number) contributed to the correct labeling. Interestingly, although both FF and BP took into account all three regions, they showed higher activation values in distinct areas. Similarly, expressive regions can be identified for the digits 7 and 9 of Fig. 4c,d. CAMs also showed that the different layers of the FF trained CNN provided similar, but distinct information for the classification task. For example, when considering digit 7 (Appendix E, Supplementary Materials), the second layer of the network provided more information on the inner portion of the horizontal line, while the third layer responded more to the boundaries of that horizontal line.

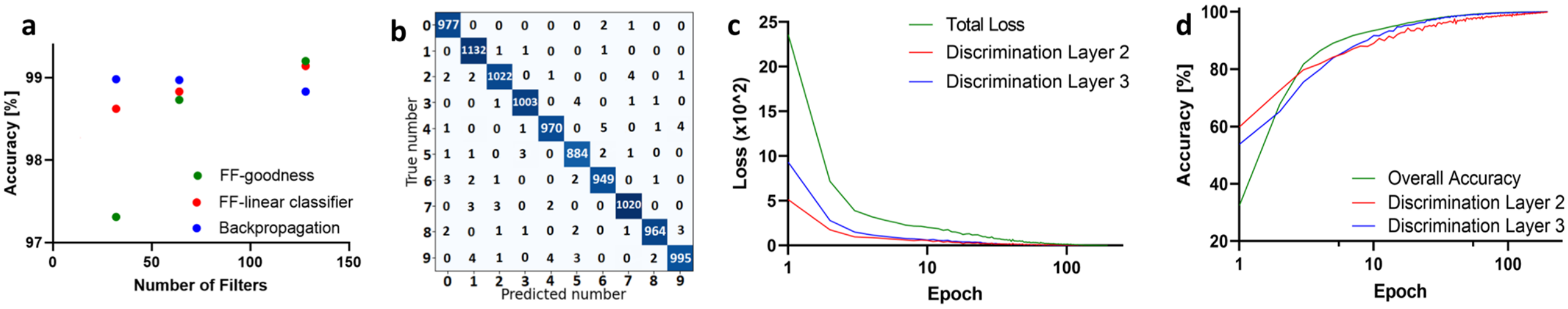

**Fig. 3**. The best MNIST performance of an FF-trained CNN architecture is comparable to the results of a backpropagation trained CNN of the same architecture. (**a**) The accuracy values obtained for CNN with three convolutional layers as a function of the number of filters in each layer, after being trained for 200 epochs with batch size 50. Filter size is 7 *times* 7, the learning rate was set to the respective optimal value of $5x10^{-5}$ for FF and $10^{-3}$ for BP. FF trained networks used labels from set 1 and a label intensity K of 35%. The values reported for BP and FF are gathered from the validation data. The green data points shows the results related to the FF trained network, with inference using the goodness comparison. In this scenario, 99.16± 0.02% accuracy was achieved with 128 filters per layer using the test data as shown by the corresponding confusion matrix reported in (**b**). (**c**) shows the loss computed for the discrimination between positive and negative training data for each hidden layer contributing to the training (red and blue lines), and the combined loss used during training (green line). (**d**) displays the discrimination accuracy of the same hidden layers (red and blue lines), and the total accuracy obtained during training (green line).

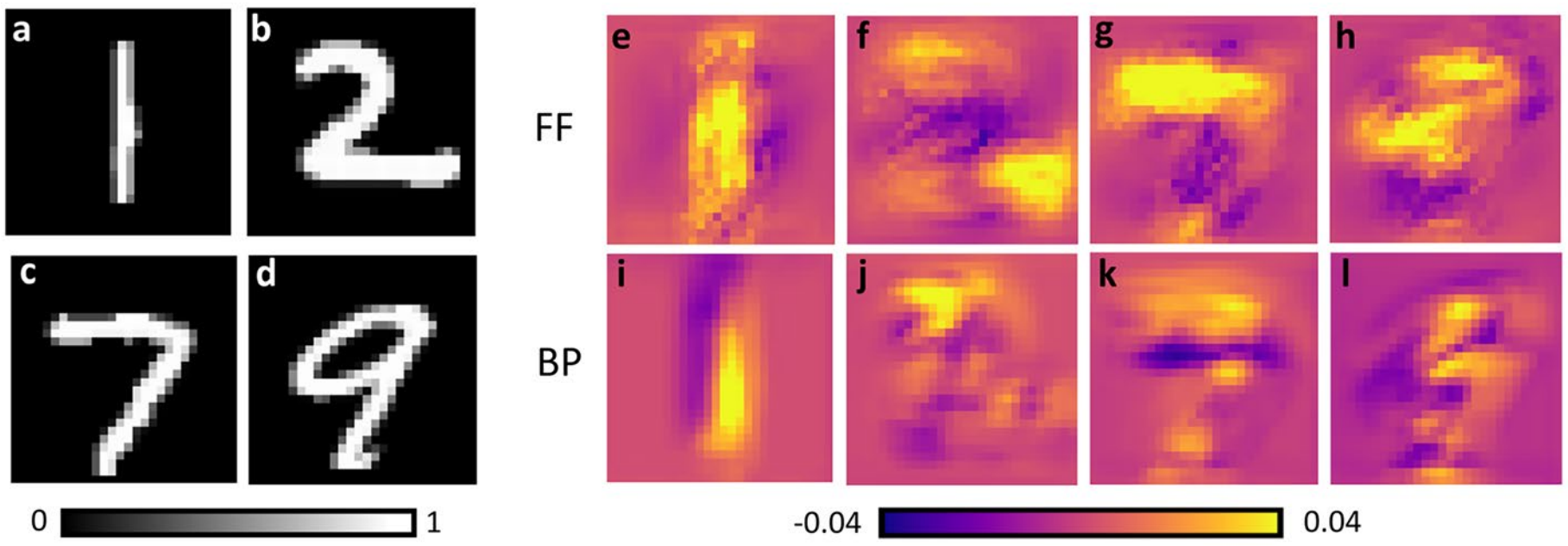

**Fig. 4**. CAMs of FF and BP trained CNNs show which image regions are considered beneficial (yellow) or deleterious (pink) by the network for making its prediction. (**a**)–(**d**) display four input images. (**e**)–(**h**) and (**i**)–(**l**) are their CAMs for FF and BP-based trainings, respectively. All examples are from a network with 16 convolutional neurons per layer, filter size $5 \times 5$, and trained with a batch size of 50 along 200 epochs. FF learning rate: $5 \times 10^{-5}$, BP learning rate: $1 \times 10^{-3}$.

## FF-trained CNNs application on CIFAR10 dataset

Although MNIST is well suited for preliminary evaluations of novel architectures, its limited morphological complexity makes it insufficient as a benchmark to fully assess algorithmic performance. CIFAR10 provides a more appropriate challenge to test both the strengths and limitations of FF training. Previous work has proposed optimizations to extend FF training to more complex datasets, including (i) deeper networks of up to 16 layers[43], (ii) convolutional group channels for positive/negative sampling[44], (iii) grouped layer training methods such as Overlapping Local Updates (OLU)[39–45], and (iv) modified loss functions[30] or the replacement of layer normalization with batch normalization[39]. Since our goal was to assess the reliability of pure FF mechanisms, we avoided hybrid strategies like OLU, which partially reintroduce backpropagation by jointly updating multiple layers. Likewise, we retained spatial labeling to demonstrate that this aspect does not represent a strong limitation. Consequently, as detailed in the Methods section and Appendix B of the Supplementary Materials, we trained a six-layer CNN with batch normalization applied before each convolution. We trained the network with 100 epochs, a batch size of 32 elements, a constant learning rate of $10^{-4}$ and a label intensity K of 30%. With backpropagation, the same CNN reached an accuracy of 85.4 ±0.4% on the test set. By applying the FF algorithm and the Fourier-based labeling, we obtained a classification accuracy of 60.9 ±0.6% on the test set, while the morphology-based labeling led to an accuracy of the 68.6 ±0.5%. Layer-wise binary accuracy values and CAMs shown in Fig. 5 suggest a mechanistic explanation. When Fourier labels are used, the injected patterns are too simple if compared to the images morphological variability. Therefore, the network finds an easier optimization path by detecting the label pattern as a shortcut to discriminate positive and negative samples. This behaviour confines discriminative signal in the earlier layers and reduces the need for deeper layers to learn robust morphological features: training converges to these shortcut solutions rather than exploring complex structures. A comparison with other approaches, reported in Appendix F of Supplementary Materials, show that our results are comparable with the state of the art of pure FF approaches.

## FF-trained CNNs application on CIFAR100 dataset

To evaluate the feasibility of spatially-extended labeling with a large number of classes, we applied the FF algorithm to the CIFAR100 dataset. CIFAR100 keeps a level of morphological complexity comparable to CIFAR10, while introducing the additional challenge of discriminating among ten times more classes. For Fourier-based spatially-extended labels, we generated 2000 unique candidate sets of frequencies and orientations. From these, we selected the 100 sets that minimized the maximal internal correlation within each group, thereby ensuring minimal internal redundancy and maximal discriminative power among the patterns. Similarly, for morphology-based labels, we randomly sampled 2000 candidate labels generated from combinations of simple transformations and their parameters. Again, we considered the 100 labels that minimized the maximal internal correlation. We trained the same CNN architecture we used for CIFAR10, with the same hyperparameters and for 100 epochs. With optimized label sets, we achieved test accuracies of 37.4 ± 0.5% (Fourier) and 38.2 ± 0.5% (morphology). Randomly sampled labels from the 2000 candidates yielded lower accuracies of 37.0 ± 0.7% and 36.5 ± 0.6%, respectively. Although prior work on FF training with CIFAR100 is limited (see Appendix G), our results are comparable with existing approaches. These results show that our labels can scale up to 100 classes if label diversity is carefully optimized. Fourier-based labels are still limited by filters or image size, whereas morphology-based alternative offers greater versatility and higher accuracy. Feasibility is further supported by the binary classification performance of each layer during the training phase (Appendix H): failure to distinguish positive from negative samples would have suggested a breakdown of the labeling mechanism.

## Discussion

CNNs are considered the gold standard in deep learning-based image analysis. In biomedical imaging, they overcome the drawbacks of subjective analysis in semiquantitative visual inspection of samples[46], and they support experts during their daily clinical routine by reducing their workload[47]. Furthermore, their exploitation of the spatial information within images makes them suitable for the deployment of explainable AI tools (such as class activation maps), which highlight the image regions that contribute most significantly to the classification outcome. Our results were obtained without implementing all the possible and suggested optimizations such as enforcing the symmetry of the loss function[30] or choosing hard, i.e. easily confused, labels for the negative dataset, as suggested by Hinton[5]. An open question remains if this technique will supersede BP in specific applications. We believe that this potential exists, especially in the cases of neuromorphic hardware and unsupervised learning. However, a better understanding of the FF training will also expand our understanding of the generic concept of neuronal information processing in all its breadth from biological systems to reservoir computing. The demonstrated ability to implement class activation maps offers an initial insight into these research topics. Achieving deeper insights will also mean understanding how the two innovations of FF, providing positive and negative labels and computing a locally defined goodness parameter, contribute to its success individually and synergetically[33]. Our comparison of Fourier and morphology-based labels already highlights how label design can dominate or facilitate feature learning. A complementary analysis of the goodness function, in turn, would clarify its role in shaping layer-wise representations. In this work we deliberately followed the original formulation of the FF algorithm, which relies on explicit positive and negative examples. While recent studies have questioned the necessity of negatives and proposed variants without them, there is increasing evidence that negative data can still be advantageous in several contexts. In particular, hard negatives are known to sharpen class boundaries and improve feature discrimination, especially in fine-grained classification tasks. In the self-supervised learning literature, contrastive approaches such as SimCLR or MoCo depend critically on negatives, and even more recent methods like AdCo demonstrate that learnable negatives can boost efficiency and stability with a relatively small memory bank. Moreover, in scenarios with class imbalance or implicit feedback (e.g., recommender systems), carefully sampled negatives help reduce bias and improve generalization. For these reasons, investigating how FF handles negative data and how their generation can be optimized remains a promising direction, particularly in unsupervised or contrastive learning settings. Moreover, a better understanding why it is beneficial to exclude the first layer during the goodness computation (cf. Appendix B of Supplementary Materials) would be desirable. Subsequent work on FF training should also address its ability to train even deeper networks, most likely expanding preliminary studies which already showed successful convergence in FF-trained network up to 16 layers[31,39,43]. Similarly, advances in optimizing the goodness-based inference scheme, which presently requires multiple forward passes and is therefore computationally costly, would greatly improve the practicality of FF-based networks for real-world applications with many output classes. From a theoretical perspective, while a formal proof of convergence is still lacking, the complementary feature representations uncovered by CAMs indicate that different layers settle into distinct states. At the same time, our comparison of Fourier- and morphology-based labels shows that FF dynamics can converge either to trivial shortcut solutions or to richer morphological features, highlighting the possibility of regime shifts reminiscent of bifurcations in dynamical systems. Although we did not observe signs of chaotic oscillations, these findings suggest that FF training in feature space is highly sensitive to the structure of the labels, and that further theoretical analysis is needed to rigorously establish stability, convergence conditions, and the presence or absence of collapse. From an applicative perspective, we have already explored the ability of FF training to work with larger and more complex datasets, in particular concerning the bioimaging field, showing its ability to extract different features, compared to the standard backpropagation algorithm[38]. This evidence, here shown with CAMs by the different handling of the features between FF and BP, could pave the way for a more complete unveiling of the complexity of the information provided by bioimaging datasets: potential implementations of FF-trained unsupervised architectures and novel analysis methods that take advantage of FF and BP algorithms to extract a greater number of useful features are viable paths. Beyond learning rules, biological plausibility also involves modeling the physical processes of neurons and their connections. Recent neuromorphic and theoretical studies have incorporated spiking activity, axonal conduction delays, and even simplified models of myelination to simulate the temporal dynamics of action potentials and the efficiency of signal propagation in biological circuits. Similarly, dendritic computation models emphasize the role of local, non-linear integration within single neurons. These developments illustrate that plausibility requires both local learning mechanisms and realistic neuronal dynamics. In this broader context, the FF algorithm offers a complementary perspective: it introduces a local, phase-based update rule that could in principle be combined with spiking neuron models[27] or with architectures that explicitly account for conduction and compartmental dynamics. Such integration could represent an important step toward bridging abstract algorithmic efficiency with biophysically grounded simulations of the brain.

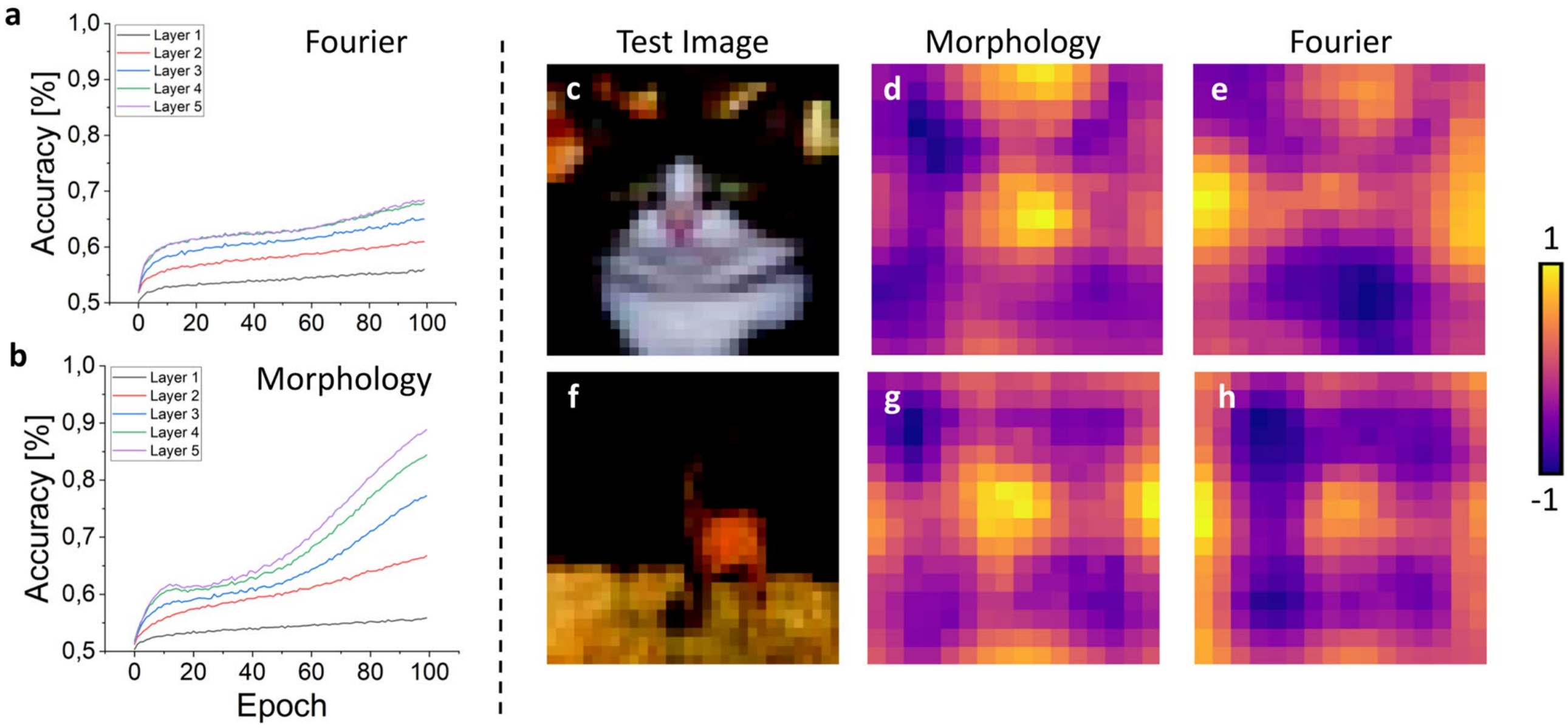

**Fig. 5**. The choice of a different spatial labeling affect the learning process. (**a**) and (**b**) show the binary accuracy values (capability of discriminating between positive and negative dataset) of the hidden layers of the FF-trained CNN tested on the CIFAR10 dataset. While the first layers, focused on simpler and rougher image features, behave similarly under both labeling strategies, deeper layers benefit from morphology-based labels but are hindered when labels are simple patterns. Given the same dataset images (**c**) and (**f**), the normalized CAMs related to the morphology-based labeling (**d**) and (**g**) more localized activation, Fourier-based labeling yields maxima confined largely to image borders (**e**, **h**).

## Data availability

The underlying code and datasets supporting the findings of this study are available upon request in the Zenodo repository at DOI:10.5281/zenodo.11571949 [40]. All data generated or analyzed during the study are included in the same repository.

## Acknowledgements
The authors also thank Christian Dullin for insightful discussions. The authors gratefully acknowledge the computing time granted by the Resource Allocation Board and provided on the supercomputer Grete at NHR-Nord@Göttingen as part of the NHR infrastructure.

## Author contributions
R.S. and M.S. ideated the FF-trained CNN approach, conceived the experiments and analyzed the results. R.S. wrote the python code. R.S., M.S. and F.A. wrote the manuscript. R.S. and A.K, conducted the experiments, M.S. and F.A. supervised the project, R.S. provided research computing resources and F.A. provided research fundings.

## Funding
Open Access funding enabled and organized by Projekt DEAL. This project has received funding from the Innovative Medicines Initiative 2 Joint Undertaking (JU) under grant agreement number 101034427. The JU receives support from the European Union's Horizon 2020 research and innovation program and EFPIA. The JU is not participating as a contracting authority in this procurement. This project was also funded by the Ministry for Science and Culture of Lower Saxony as part of the project "Agile, bio-inspired architectures" (ABA). Preliminary calculations for this research were conducted with computing resources under the NHR-Verein project nib00048.

## Declarations

## Competing interests
The authors declare no competing interests.

## Additional information
**Supplementary Information** The online version contains supplementary material available at https://doi.org/10.1038/s41598-025-26235-2.

**Correspondence** and requests for materials should be addressed to R.S.

**Reprints and permissions information** is available at www.nature.com/reprints.

**Publisher's note** Springer Nature remains neutral with regard to jurisdictional claims in published maps and institutional affiliations.